\documentclass[letterpaper, 10 pt, conference]{ieeeconf}  % Comment this line out if you need a4paper

\IEEEoverridecommandlockouts                              % This command is only needed if 
\usepackage{times}
\usepackage{graphicx}
\usepackage{amsmath,amssymb,amsopn,amstext,amsfonts}
\usepackage{cancel}
\usepackage[space]{cite}
\usepackage{pdfsync}
\usepackage{balance}
\usepackage{color}
\usepackage{mathtools}
\usepackage{algpseudocode}
\usepackage{algorithm}
\usepackage{bm}

\usepackage{diagbox}
\usepackage{float}
\usepackage{epstopdf}
\usepackage{pifont}
\usepackage{fixltx2e}
\usepackage{amsmath}
\usepackage{multirow}
\usepackage{url}
\usepackage{verbatim}
\usepackage{amssymb}
\makeatletter
\let\NAT@parse\undefined
\makeatother
\usepackage[linkcolor=black,citecolor=black,urlcolor=black,colorlinks=TRUE]{hyperref}
\usepackage{booktabs}
\usepackage{graphicx}
\usepackage{subcaption}
\usepackage{gensymb}
\usepackage{stfloats}
\usepackage{amsmath}
\usepackage{nccmath}
\usepackage{arydshln}
\usepackage{siunitx}
\usepackage{tabularx} % 用于 X 列
\usepackage{array}    % 用于 >{\centering}
\graphicspath{{./FIGURES/}}
\DeclareGraphicsExtensions{.png,.jpg,.eps,.pdf}
\usepackage{booktabs} % 必须：用于 \toprule, \midrule
\usepackage{amsmath}  % 必须：用于 \text, \boldsymbol
\usepackage{amssymb}  % 必须：用于 \mathcal
\usepackage{graphicx} % 必须：用于 \resizebox (调整表格大小)
\usepackage{amssymb}
\usepackage{xurl}
\title{\LARGE \bf
 Flying in Clutter on Monocular RGB by Learning in 3D Radiance Fields with Domain Adaptation
}

\author{ 
	Xijie Huang\textsuperscript{1,2},
    Jinhan Li\textsuperscript{1,2},
	Tianyue Wu\textsuperscript{1,2},
    Xin Zhou\textsuperscript{2}\\
    Zhichao Han\textsuperscript{1,2,\textdagger},
     and Fei Gao\textsuperscript{1,2,\textdagger}
	\thanks{
\textsuperscript{\textdagger}\emph{Corresponding authors: Zhichao Han and Fei Gao}
	} 
	\thanks{\textsuperscript{1}State Key Laboratory of Industrial Control Technology, Zhejiang University, Hangzhou 310027, China.}
    \thanks{\textsuperscript{2}Differential Robotics,  Hangzhou 311121, China.}
	\thanks{E-mail:{\tt\small \{xijiehuang, fgaoaa\}@zju.edu.cn}}
}

\begin{document}
    	
    % \makeatletter
    % \let\@oldmaketitle\@maketitle
    % \renewcommand{\@maketitle}{\@oldmaketitle
    % 	% 删除了嵌入的图片和标题代码
    % }
    % \makeatother
    % \maketitle
    % \setcounter{figure}{1}
    % \thispagestyle{empty}
    % \pagestyle{empty}

	\makeatletter
	\let\@oldmaketitle\@maketitle
	\renewcommand{\@maketitle}{\@oldmaketitle
		% \begin{center}
			\includegraphics[width=1.0\linewidth]{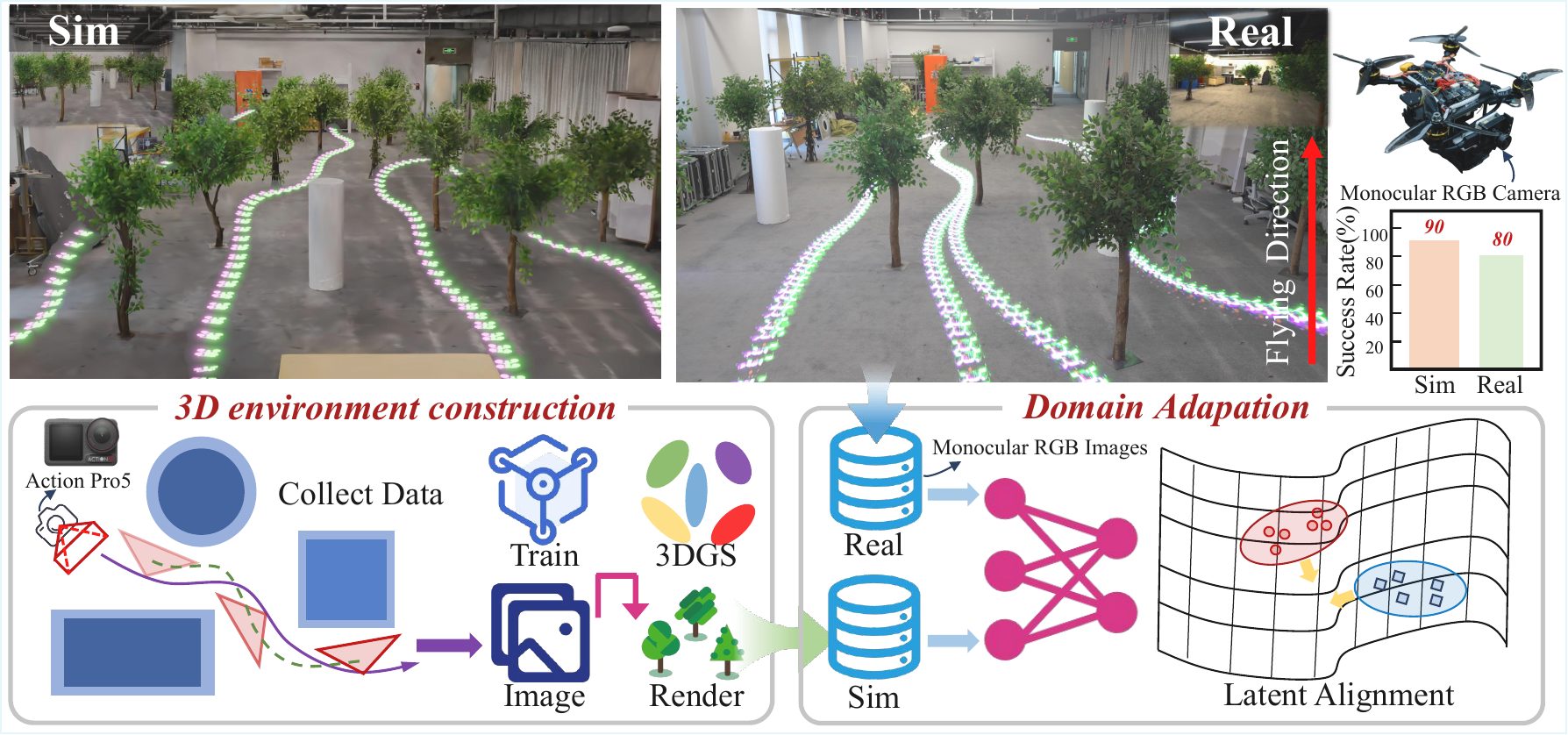}
		% \end{center}
		\captionsetup{font={small}}
		\captionof{figure}{
			\label{fig:top}
\textbf{Overview of our RGB-based navigation framework.} The top panels demonstrate that our method effectively mitigates the sim-to-real gap. The bottom panel illustrates the pipeline of 3D environment construction and domain adaptation.
% The left panel displays the real-world trajectory and the right panel shows the 3DGS-rendered trajectory. While the success rate reached 100\% in simulation, we achieved an 80\% success rate in the real world. The results demonstrate our system can reproduce the success from simulation to
% reality using raw RGB images.
		}
  	}
	\makeatother
	\maketitle
        	\setcounter{figure}{1}
	\thispagestyle{empty}
	\pagestyle{empty}

\begin{abstract}

    Modern autonomous navigation systems predominantly rely on lidar and depth cameras. However, a fundamental question remains: Can flying robots navigate in clutter using solely monocular RGB images? Given the prohibitive costs of real-world data collection, learning policies in simulation offers a promising path. Yet, deploying such policies directly in the physical world is hindered by the significant sim-to-real perception gap. Thus, we propose a framework that couples the photorealism of 3D Gaussian Splatting (3DGS) environments with Adversarial Domain Adaptation. By training in high-fidelity simulation while explicitly minimizing feature discrepancy, our method ensures the policy relies on domain-invariant cues. Experimental results demonstrate that our policy achieves robust zero-shot transfer to the physical world, enabling safe and agile flight in unstructured environments with varying illumination.
 
\end{abstract}

\section{Introduction}

For Unmanned Aerial Vehicles (UAVs), effective environmental understanding is the core of safe and robust navigation. Within the realm of perception systems, unlike depth sensors or lidar, which necessitate precise extrinsic calibration across multiple components \cite{neuralfeels}, monocular RGB cameras represent the most compact and natural perception modality. Offering distinct advantages in size, weight, and power, they provide a promising sensing solution for miniature platforms and non-rigid UAV structures where hardware constraints are strict. However, existing approaches have yet to fully exploit this potential. Traditional methods, which decouple navigation into localization, mapping, and trajectory optimization \cite{srswarm}, struggle to extract reliable geometric structures from monocular imagery. While data-driven approaches offer a compelling end-to-end alternative \cite{geles2024demonstrating,yuan_gap}, the majority still depend on explicit geometric priors provided by lidar \cite{lidar} or depth sensors \cite{backnewton}. Thus, a question arises: Is it feasible for flying robots to achieve robust autonomous navigation relying on monocular RGB information?

While collecting real-world data is costly, training in simulation serves as a more efficient alternative paradigm. In practice, learning navigation policies directly from raw monocular images in simulation presents two major challenges. First, unlike depth maps or point clouds, monocular RGB data is inherently implicit and scale-ambiguous \cite{cad2rl}, making the extraction of implicit geometric structures and navigation information difficult to model via traditional rule-based methods. Second, the sim-to-real gap is particularly pronounced in the visual modality, which severely hinders the transfer of control policies learned entirely from simulation rollouts. The complex illumination variations, diverse textures, and sensor noise ubiquitous in the physical world often induce a significant visual distribution shift, causing policies trained in simulation to degrade or fail upon real-world deployment.
Existing solutions typically resort to explicit intermediate representations, such as optical flow \cite{opticalflow}, to bypass the geometric ambiguity, or employ visual domain randomization \cite{cad2rl} to mitigate the distribution shift. Unfortunately, the former inevitably introduces modular latency and loses rich semantic information, while the latter relies on unrealistic augmentations that compromise performance.

% \cite{visualdomain,domainsurvey}.

% % Prior vision-based navigation approaches predominantly relied on explicit intermediate representations, such as optical flow \cite{opticalflow}, to extract environmental cues.

In this paper, we propose a navigation framework designed to tackle these specific challenges. First, to address the difficulty of modeling implicit geometric structures, we leverage end-to-end reinforcement learning (RL) to learn policies directly from raw RGB images. This data-driven paradigm empowers the agent to implicitly extract reliable navigation cues from photometric inputs. Second, to overcome the visual sim-to-real gap, we introduce a high-fidelity simulation pipeline combined with domain adaptation (DA). We employ 3D Gaussian Splatting (3DGS) \cite{kerbl3Dgaussians} to reconstruct real-world scenes, establishing a highly photorealistic training environment. To mitigate the prohibitive computational cost of standard 3DGS during RL training, we accelerate the pipeline via model pruning and parallelized backends \cite{ye2025gsplat}. These optimizations enable real-time rendering with a peak throughput of approximately 30,000 frames per second. Furthermore, inspired by \cite{depthtransfer}, we explicitly integrate an adversarial domain adaptation mechanism \cite{dann} to align feature distributions between simulation and reality. Ultimately, our method achieves successful zero-shot transfer, demonstrating robust navigation capabilities in real-world scenarios with randomly distributed obstacles and varying illumination.

Overall, the main contributions of this paper can be summarized as follows:
\begin{itemize}	
	\item [1)]
We propose an end-to-end training paradigm for navigation using monocular RGB inputs.
	\item [2)]
We leverage model pruning and parallelized backends to accelerate 3DGS rendering.
	\item [3)] 
We utilize domain adaptation to bridge the gap between the 3DGS simulation domain and the UAV camera domain, facilitating robust sim-to-real transfer.
	\item [4)] 
We validate through simulation and real-world experiments that our pipeline enables safe and agile autonomous flight. 
\end{itemize}

\section{Related Work}

\subsection{Autonomous Navigation with  Geometric Sensors}
\label{sec:geo_nav}

Significant progress has been made in autonomous UAV navigation by leveraging sensors that provide explicit geometric information.
Loquercio et al. \cite{learninwild} utilized imitation learning to map depth images directly to control actions, successfully demonstrating high-speed flight in wild environments.
Subsequently, Zhang et al. \cite{backnewton} exploited a differentiable simulator to provide first-order gradient information for efficient policy updates, achieving robust obstacle avoidance via reinforcement learning based on depth inputs.
Parallel to depth-based methods, other works have focused on processing high-dimensional point clouds. For instance, a recent work \cite{lidar} proposed a method to handle raw lidar data, navigating safely around thin obstacles.

However, these approaches rely on lidar and depth cameras, which are generally more expensive, heavier, and more energy-consuming. Furthermore, in scenarios demanding semantic understanding of the environment, such purely geometric methods may prove suboptimal.

\subsection{Visual Navigation from Monocular RGB}
\label{sec:rgb_nav}

To address the limitations of geometric sensors, research has shifted towards navigation based on monocular RGB inputs.
Early works \cite{singla2019memory, s17051061} utilized reinforcement learning to train collision avoidance policies within simulation. Notably, CAD$^2$RL \cite{cad2rl} introduced a sim-to-real paradigm, leveraging extensive visual domain randomization to achieve successful transfer to the physical world. However, the limited visual fidelity of simulators failed to provide sufficiently realistic training data. This discrepancy widened the sim-to-real gap and degraded policy performance in real-world scenarios.

Another prevailing paradigm involves extracting explicit geometric features from RGB images before feeding them into the control policy. Methods utilizing depth estimation \cite{depthanything} or optical flow estimation \cite{neuralflow} employ these features as intermediate representations to simplify the learning task. While effective, these additional processing modules introduce computational cost and inference latency \cite{opticalflow}. Moreover, relying on such intermediate representations inevitably leads to information loss, thus limiting the performance of high-level tasks.

With the recent advent of 3D Gaussian Splatting (3DGS) for high-fidelity scene reconstruction, new possibilities for realistic training have emerged.
GraD-Nav \cite{gradnav} proposed training navigation policies within photorealistic 3DGS environments. Despite its success in simulation, the method is primarily designed for simple gate-traversal tasks and relies on hand-crafted reward shaping (e.g., trajectory waypoints). This dependence constrains the agent's exploration capabilities, often leading to sub-optimal policies.
% In contrast, our approach leverages an effective sim-to-real framework that enables agile obstacle avoidance in complex environments without the need for reference trajectories or complex reward engineering.

\section{Training Pipeline for RGB Navigation Task}
\subsection{3DGS Formulation and Environment Integration}

The original 3D Gaussian Splatting method represents the scene as a set of 3D Gaussian primitives, each parameterized by a spatial mean, a full covariance matrix, view-dependent radiance, and an opacity value. During rendering, the color of each pixel is obtained by front-to-back volumetric compositing of all Gaussians whose projected footprints cover that pixel:
\begin{equation}
    C = \sum_{i \in \mathcal{N}} c_i \alpha_i \prod_{j=1}^{i-1} (1 - \alpha_j),
\end{equation}
where $\mathcal{N}$ denotes the set of Gaussians intersecting the pixel, $c_i$ is the view-dependent color (radiance) of Gaussian $i$, and $\alpha_i \in [0,1]$ is its effective opacity at that pixel.

However, reinforcement learning typically requires large batch sizes during training. To satisfy this requirement, we adapt 3DGS pruning introduced in Speedy Splat \cite{HansonSpeedy}. For each rendered Gaussian, its projection onto the 2D image plane is an ellipse. The Gaussian-to-tile mappings are constrained strictly to the tiles that overlap with the projected ellipse, rather than the square bounding region employed in the original 3DGS framework (derived from a radius $r = \lceil 3\sqrt{\lambda_{max}} \rceil$, where $\lambda_{max}$ represents the maximum eigenvalue of the projected 2D covariance matrix).

To sparsify primitives, each Gaussian ($G_i$) is assigned a lightweight importance score 
\begin{equation}
    \tilde{U}_i = (\nabla_{g_i} I_g)^2
\end{equation}

where $I_g$ is the final rendered image and $g_i$ is the 2D projected scalar value of the $i$-th Gaussian ($G_i$). During training, Gaussians are pruned according to this score.

\begin{figure}[h]
	\centering
	\includegraphics[width=1\linewidth]{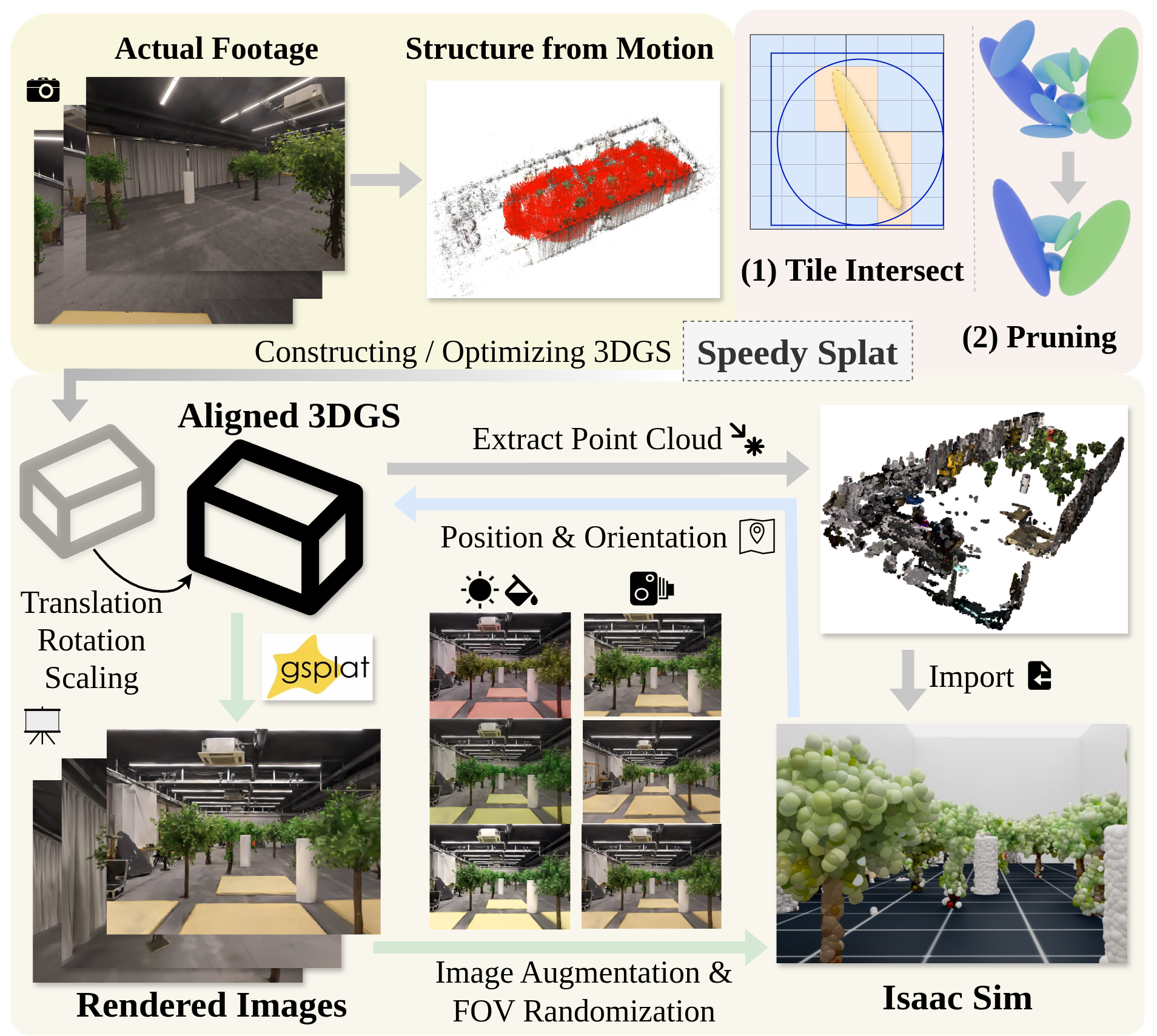}
	\captionsetup{font={small}}
	\caption{\textbf{Pipeline for constructing the 3DGS-based simulation environment.} To accelerate rendering, we employ Speedy Splat for model pruning and utilize gsplat as a parallelized rasterization backend. The aligned point cloud maps are then imported into Isaac Sim to enable depth rendering and collision detection.}
	\vspace{-0.5cm}
	\label{fig:3dgs}
\end{figure}

We constructed 10 distinct scenes using video scanning, featuring the rearrangement of obstacles (e.g., adding, removing, or repositioning) for each scene. As shown in Fig. \ref{fig:3dgs}, for each video, we performed a COLMAP Structure from Motion (SfM) \cite{schoenberger2016sfm} reconstruction, which was subsequently processed by Speedy Splat to generate a corresponding 3DGS model. The dataset was partitioned into 9 scenes for training and 1 for evaluation.

Each scene's scale and coordinate system were then transformed to align with the simulation world, and the point cloud geometry corresponding to each scene was extracted into Isaac Sim to render depth and detect collisions. In contrast, the RGB observation is rendered in real-time from the drone's current position and orientation using the trained 3DGS model. This rendering is accelerated by the gsplat library \cite{ye2025gsplat}, which utilizes vectorized, GPU-batched operations for all Gaussian projection, sorting, and compositing steps. Our pipeline achieves a peak rendering speed of 30,000 frames per second.

\subsection{Learning the policy}

\subsubsection{Actor Policy}
% The network takes RGB images ($I_{rgb}$) of size $60 \times 80$, rendered in real-time by 3DGS, as input. We first employ a normalization module to standardize the observations, followed by a three-layer CNN to extract features from $I_{rgb}$. The resulting feature map is flattened into a 4480-dimensional latent vector, $\mathbf{z}_{rgb}$. This visual feature vector $\mathbf{z}_{rgb}$ is concatenated with a 20-dimensional UAV state vector $\mathbf{s}_t$. This observation state is defined as:
% \begin{equation}
%     \mathbf{s}_t = [\mathbf{v}_b, \mathbf{R}, \mathbf{d}_{g}, z_c, z_t, \mathbf{a}_{last}]
% \end{equation}
% where $\mathbf{v}_b \in \mathbb{R}^3$ is the velocity in the body frame, $\mathbf{R} \in \mathbb{R}^9$ is the rotation matrix, $\mathbf{d}_{g} \in \mathbb{R}^3$ is the normalized goal vector in the body frame, $z_c, z_t \in \mathbb{R}$ are the current and target heights, and $\mathbf{a}_{last} \in \mathbb{R}^3$ represents the last actions. The combined $4480+20$ dimensional representation are processed by a GRU with a hidden dimension of 512. The GRU's output is passed through an MLP, and a final projection layer maps the features to the 4-dimensional normalized action space $\mathbf{a}_t = [T, \phi, \theta, \psi]$. These represent the desired thrust, roll, pitch, and yaw, respectively, which are subsequently converted to torque and thrust commands via a cascaded PID controller.
The network accepts two inputs: RGB images ($I_{rgb}$) of size $60 \times 80$ rendered in real-time by 3DGS, and a 20-dimensional UAV state vector $\mathbf{s}_t$. The state vector is defined as:
\begin{equation}
    \mathbf{s}_t = [\mathbf{v}_b, \mathbf{R}, \mathbf{d}_{g}, z_c, z_t, \mathbf{a}_{last}]
\end{equation}
where $\mathbf{v}_b \in \mathbb{R}^3$ denotes the body-frame velocity, $\mathbf{R} \in \mathbb{R}^9$ is the rotation matrix, $\mathbf{d}_{g} \in \mathbb{R}^3$ is the normalized goal vector, $z_c, z_t \in \mathbb{R}$ are current and target heights, and $\mathbf{a}_{last} $ corresponds to the previous action. We first employ a normalization module to standardize both input observations ($I_{rgb}$ and $\mathbf{s}_t$). Subsequently, a three-layer CNN extracts features from the normalized images, flattening them into a 4480-dimensional latent vector, $\mathbf{z}_{rgb}$. This visual embedding is then concatenated with the normalized state vector $\mathbf{s}_t$. The combined representation is processed by a GRU with 512 hidden units to capture temporal dynamics. Finally, the output is passed through an MLP and a projection layer to produce the normalized action $\mathbf{a}_t = [T, \phi, \theta, \psi]$, which is converted to torque and thrust commands via a cascaded PID controller.
\subsubsection{Critic Policy}
The Critic network shares a similar architecture to the Actor. To stabilize the training of the value function, we leverage the Asymmetric Actor-Critic paradigm, which utilizes privileged information available in simulation \cite{asymmetric}. Specifically, in addition to the standard actor observation ($I_{rgb}$), the Critic's CNN input is augmented with a $60 \times 80$ depth map ($I_d$) as auxiliary feature information from Isaac Sim. The effectiveness of incorporating this privileged depth information in stabilizing the training will be validated in Section \ref{sec:ablation}.
% \subsubsection{Low-Level Controller}
% A cascaded P/PID controller, similar to PX4, was designed to calculate the desired collective thrusts $F_{des}$ (from the policy) and the desired body-frame torques $\boldsymbol{\tau}_{des}$. The outer loop receives the desired angle from policy and outputs the desired bodyrate to the inner loop. Then, a standard inverse dynamics equation is used to calculate the desired torque $\boldsymbol{\tau}_{des}$:
% \begin{equation}
%  \boldsymbol{\tau}_{des} = \mathbf{J}\dot{\boldsymbol{\omega}}_{des} + \boldsymbol{\omega} \times (\mathbf{J}\boldsymbol{\omega})
% \end{equation}
% where $\mathbf{J}$ is the diagonal inertia matrix and $\dot{\boldsymbol{\omega}}_{des}$ is the desired angular acceleration from the inner loop.
\subsubsection{Termination}
An episode terminates when (i) the UAV's altitude (z-axis position) falls outside the predefined safety range of [0.1m, 1.7m], (ii) a collision with the environment is detected, which is checked in real-time using a global, inflated point cloud occupancy map, or (iii) the UAV successfully reaches the goal. When the agent meets any termination condition, it is reset to a new starting point randomly sampled from an initial free-space region.

\begin{figure*}[h]
	\centering
	\includegraphics[width=1.0\linewidth]{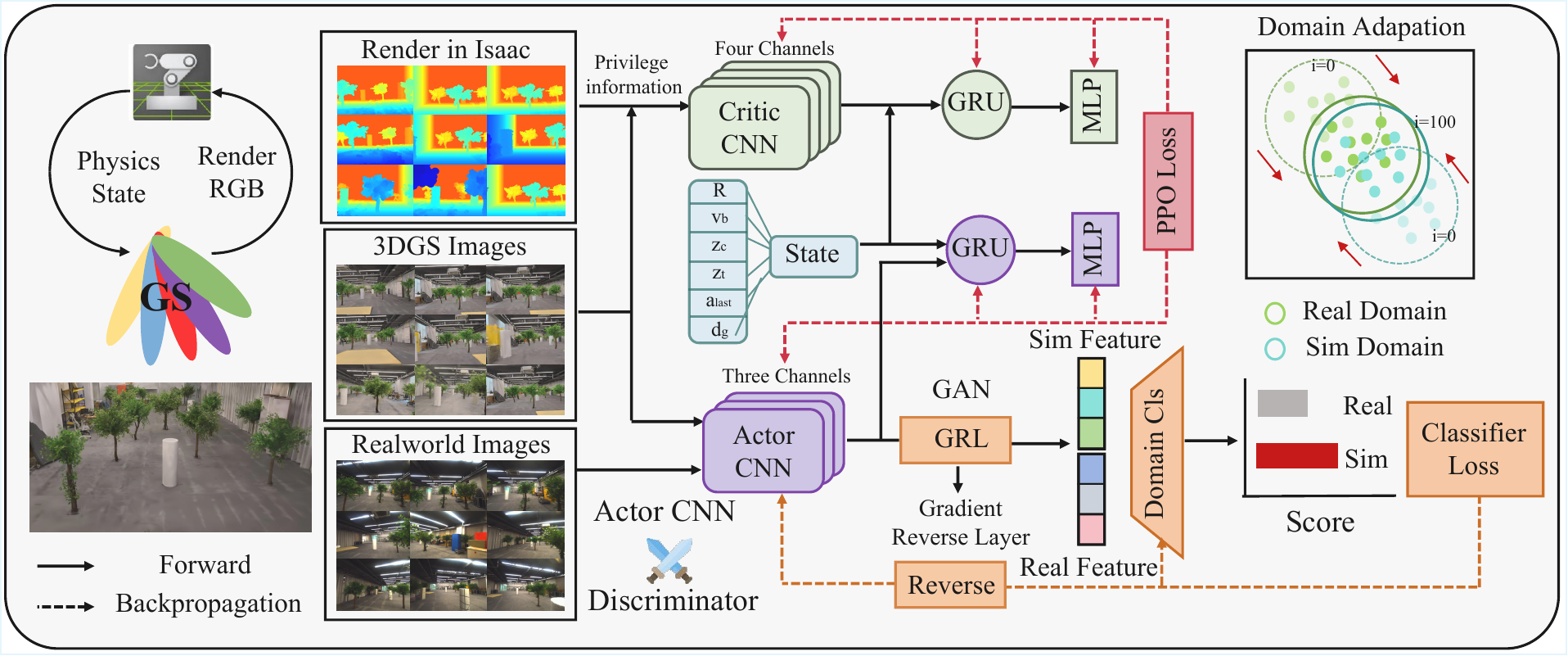}
	\captionsetup{font={small}}
	\caption{
      \textbf{Overview of the proposed RL training framework.} The architecture features an asymmetric actor-critic structure and employs an adversarial domain adaptation module to bridge the sim-to-real gap.
	}
	\vspace{-0.5cm}
	\label{fig:trainpipeline}
\end{figure*}

\subsubsection{Reward Function}
To achieve fast and stable UAV navigation in cluttered environments, we design a composite reward function. The total reward $r_t$ at each timestep $t$ is a weighted sum of several components:
\begin{equation}
    r_t = \sum_{i} w_i r_i
\end{equation}
where $r_i$ is an individual reward component and $w_i$ is its corresponding weight.

For the Distance Reward, we use the change in goal distance $r_{\text{distance}} = d_{t-1} - d_t$, where $d_t$ is the Euclidean distance to the goal. This reward is clamped per-step to the maximum possible distance change, $d_{\text{step}} = v_{\max} \cdot \Delta t$. The components are detailed in Table \ref{tab:reward_function}.

\begin{table}[h]
\centering
\caption{Reward function components and their respective weights. $v_{\max}$ is set to 2 m/s in this experiment.}
\label{tab:reward_function}
\begin{tabularx}{\columnwidth}{ l >{\centering\arraybackslash}X c }
\toprule
\textbf{Reward} & \textbf{Equation ($r_i$)} & \textbf{Weight ($w_i$)} \\
\midrule
Obstacle Collision & $r_{\text{collision}}$ & -80.0 \\

Z-Velocity Penalty & $-\min(|v_{b,z}|, 1.0)$ & -0.5 \\
Action Magnitude & $\|\mathbf{a}_{t, 0:2}\|^2$ & -0.3 \\
Action Change & $\|\mathbf{a}_{t, 0:2} - \mathbf{a}_{t-1, 0:2}\|^2$ & -0.6 \\
Distance Reward & $\operatorname{clamp}(d_{t-1} - d_t, -d_{\text{step}}, d_{\text{step}})$ & 30.0 \\
Success Reward & $r_{\text{success}}$ & 80.0 \\
Velocity Excess &
    \begin{tabular}{@{}c@{}} 
      $\max(-(e^{\|\mathbf{v}_t\| - v_{\max}} - 1), -5.0)$ \\
      \small $\text{if } \|\mathbf{v}_t\| > v_{\max}$, else $0$
    \end{tabular}
& -1.0 \\
\bottomrule
\end{tabularx} 
\end{table}

\section{Sim-to-real Transfer}

\subsection{Domain Adaptation}

To reconstruct high-quality 3DGS environments for training, we utilize the DJI Osmo Action 5 Pro camera\footnote{DJI Osmo Action 5 Pro: \url{https://www.dji.com/osmo-action-5-pro}}  for data recording. However, during real-world deployment, cost and payload constraints necessitate the use of affordable and lightweight sensors. In our experiments, we employ a Hikvision industrial camera\footnote{Hikvision industrial camera: \url{https://www.hikrobotics.com/cn/machinevision/productdetail/?id=9706}} for onboard perception. As illustrated in Fig. \ref{fig:trainpipeline}, this sensor mismatch introduces a significant gap which arises from discrepancies in lens distortion, color style, exposure, and intrinsic 3DGS reconstruction artifacts. 

To bridge this gap, we employ a DA strategy to align the latent feature spaces of the two domains efficiently. It needs only approximately several minutes of real-world camera video.

As shown in Fig. \ref{fig:trainpipeline}, a discriminator is designed after the Actor CNN Encoder. It aims to distinguish whether a latent feature vector comes from the 3DGS or the real world. Conversely, the encoder aims to generate domain-invariant features that can mislead the discriminator. 

During the forward pass, the Gradient Reversal Layer acts as an identity function. In backpropagation, it reverses the gradient flowing from the discriminator to the encoder, scaled by a factor $\lambda$:
\begin{equation}
    \frac{\partial \mathcal{L}_{\text{DA}}}{\partial \theta_E} = - \lambda \frac{\partial \mathcal{L}_{\text{DA}}}{\partial \theta_D}
\end{equation}
where $\theta_E$ and $\theta_D$ are the parameters of the encoder and discriminator, respectively.

The discriminator is trained using a binary cross-entropy loss to maximize  classification accuracy:
\begin{equation}
    \mathcal{L}_{\text{DA}} = - \mathbb{E}_{z_s \sim \mathcal{D}_s[\log D(z_s)]} - \mathbb{E}_{z_t \sim \mathcal{D}_t[\log (1 - D(z_t))]}
\end{equation}
where D is discriminator, $D_s$ and $D_t$ are the latent feature from 3DGS (source) and real-world (target).

In contrast to \cite{depthtransfer}, which decouples the training of the discriminator and the RL agent, we integrate the domain loss directly into the PPO training objective. This approach reduces the loss of optimality caused by fragmented training, ensuring the policy retains its navigation capabilities while effectively aligning features across domains. The total loss $\mathcal{L}_{\text{total}}$ is a weighted sum of the PPO loss and the DA loss:
\begin{equation}
    \mathcal{L}_{\text{total}} = \lambda_1 \cdot \mathcal{L}_{\text{PPO}} + \lambda_2 \cdot \mathcal{L}_{\text{DA}}
\end{equation}
where $\mathcal{L}_{\text{PPO}}$ includes the policy surrogate loss, value function loss, and entropy term. The coefficients $\lambda_1$ and $\lambda_2$ are carefully tuned to balance task performance with feature alignment, ensuring the policy learns robust obstacle avoidance behaviors and generalizes to the real-world latent space.

\begin{figure}[h]
	\centering
	\includegraphics[width=1.0\linewidth]{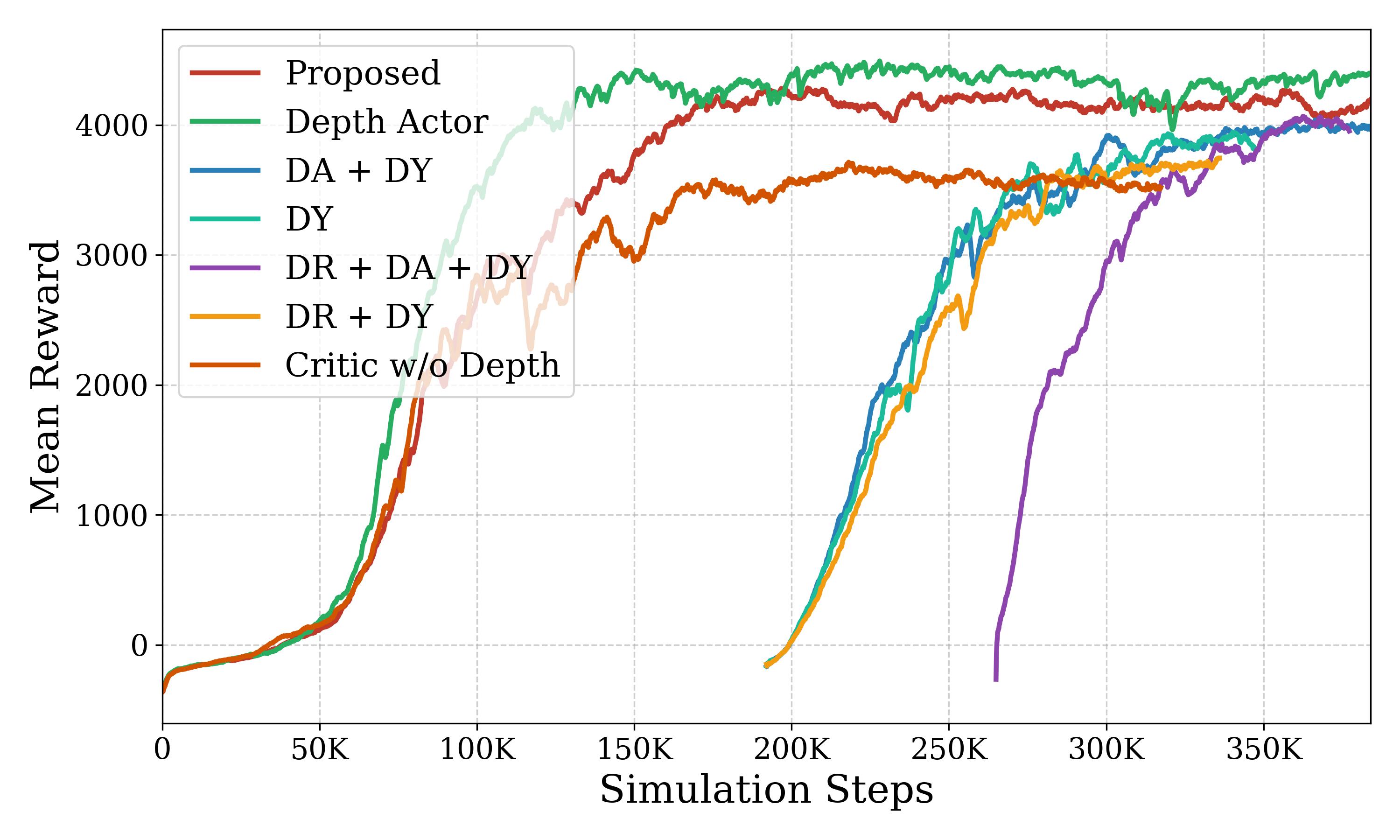}
	\captionsetup{font={small}}
	\caption{
    \textbf{Training curves for ablation studies.} We compare the mean reward convergence across different policy inputs and sim-to-real strategies. Abbreviations: DA (Domain Adaptation), DR (Visual Domain Randomization), and DY (Dynamic Domain Randomization). Note: DY is applied as a fundamental component to mitigate the dynamics gap between simulation and reality. The DA+DY, DR+DY and DY are trained based on the Proposed, while DR+DA+DY is trained based on DR+DY.
	}
	\vspace{-1.0cm}
	\label{fig:Ablation}
\end{figure}

\subsection{Domain Randomization}
To bridge the sim-to-real gap, we implement comprehensive randomization on dynamics and perception. Detailed parameters and distributions are provided in Table \ref{tab:domain_randomization}.

\subsubsection{Dynamics noise randomization}
We simulate actuator imperfections by injecting Gaussian noise and introducing variable latency into the control command, similar to \cite{teji}. Latency is modeled as a moving average over a history window. Rather than using fixed parameters, both the delay magnitude ($D$) and noise vectors are dynamically re-sampled at randomized intervals ($T$) within an episode. This stochasticity forces the policy to adapt to changing disturbances rather than overfitting to static delay patterns.

\subsubsection{Perception noise randomization}
To emulate the imperfections of real-world state estimation, we introduce noise perturbations to the UAV's linear velocity, heading vector, and altitude. Crucially, estimation errors in physical systems are rarely temporally independent. Thus, the noise dynamics are governed by Eq. \ref{eq:ou_noise}, with a mean-reversion rate of $\theta=0.1$:
\begin{equation}
\label{eq:ou_noise}
\begin{aligned}
n_{t+1} &= (1-\theta)n_t + \mathcal{N}(0, \sigma^2) \\
\hat{s}_{t+1} &= s_{t+1} + n_{t+1}
\end{aligned}
\end{equation}
where $s_{t+1}$ denotes the ground-truth state and $\hat{s}_{t+1}$ represents the policy observation.

 % by applying a linear transformation to the 0th-degree Spherical Harmonics coefficients ($C_{dc}$) to simulate diverse lighting conditions and sensor noise:

% \subsubsection{Visual  Augmentation}
Furthermore, we adapt the visual augmentation method from RoboSplat \cite{robosplat} by directly transforming the 3DGS parameters as follows:
\begin{equation}
C'_{dc} = \alpha \cdot C_{dc} + \beta + \boldsymbol{\epsilon}
\end{equation}
where $\alpha$ and $\beta$ represent scale and offset factors, respectively. Additionally, we randomize the FOV to improve generalization across obstacle scales.
% , compelling the policy to learn robust features rather than overfitting to specific shapes.

\begin{table}[h]
\centering
\caption{Domain randomization parameters and implementation details.}
\label{tab:domain_randomization}
\resizebox{\columnwidth}{!}{%
\begin{tabular}{lcc}
\toprule
\textbf{Parameter} & \textbf{Probability} & \textbf{Distribution / Value} \\ 
\midrule
\multicolumn{3}{l}{\textbf{Dynamics noise randomization}} \\
Action Noise ($\epsilon$) & 1.0 & $\sim \mathcal{N}(0, \sigma^2)$ \\
Latency Delay ($D$) & 1.0 & $\sim \mathcal{U}(0 \text{ms}, 80 \text{ms})$ \\
Resample Interval ($T$) & 1.0 & $\sim \mathcal{U}(10 \text{ms}, 100 \text{ms})$ \\

\midrule
\multicolumn{3}{l}{\textbf{Perception noise randomization}} \\
Velocity State Noise ($\sigma_{vel}$) & 1.0 & $\sim \mathcal{N}(0, 0.08)$ \\
Direction State Noise ($\sigma_{dir}$) & 1.0 & $\sim \mathcal{N}(0, 0.05)$ \\
Z-axis State Noise ($\sigma_{z}$) & 1.0 & $\sim \mathcal{N}(0, 0.03)$ \\ 
Color Scale ($\alpha$) & 1.0 & $\sim \mathcal{U}(0.8, 1.3)$ \\
Color Offset ($\beta$) & 1.0 & $\sim \mathcal{U}(-0.05, 0.05)$ \\
SH Additive Noise ($\sigma_{\text{rgb}}$) & 1.0 & $\sim \mathcal{N}(0, 0.025^2)$ \\ 
Field of View (FOV) & 1.0 & $\sim \mathcal{U}(67^{\circ}, 106^{\circ})$ \\
\bottomrule
\end{tabular}%
}
\vspace{-0.5cm}

\end{table}

\section{Evaluations}

The policy is trained using 1,024 parallel environments with a rollout length of 128 time steps per update. As illustrated in Fig. \ref{fig:Ablation}, our proposed method follows a three-stage training. It is important to note that DR refers specifically to visual DR. Since Dynamic Domain Randomization (DY) is a well-established standard for bridging the dynamics gap in sim-to-real transfer, it is applied as a fundamental setting in our real-world experiments and is not separately ablated.
In the first stage, we train a baseline policy. Subsequently, we introduce DR+DY to fine-tune the policy in the second stage. Finally, we incorporate both DR+DY+DA to align latent features in the final stage. The entire training process is completed in approximately two days on a single NVIDIA L40 GPU. For real-world deployment, the final policy is executed on an NVIDIA Jetson Orin NX, achieving an inference latency of approximately 2 ms. Real-time state estimation, such as the normalized direction vector, altitude, and velocity, is derived from an Extended Kalman Filter (EKF) that fuses optical flow and IMU measurements.
\subsection{Ablation Study}
\label{sec:ablation}
In this section, we conduct a series of ablation studies to validate the effectiveness of using privileged information (i.e., Depth) in the Critic's training and the efficacy of the rich semantic information from the RGB input.

We compare the learning curves and final success rates of four experimental setups, as shown in Fig. \ref{fig:Ablation}:
1) \textbf{Critic without depth}: Removing privileged information from the Critic; 
% 2) \textbf{Optical Flow Actor}: using Optical Flow  as the input to the Actor which estimated by \cite{neuralflow} ;
2) \textbf{Depth Actor}: A baseline using ground-truth depth for the Actor;
3) \textbf{Proposed}: Our full method (RGB-Actor, Privileged-Critic);

First, our method achieves reward levels competitive with the Depth-based baseline. It suggests that our policy successfully extracts implicit geometric features and traversability cues directly from high-dimensional photometric data. This result validates the feasibility of achieving robust navigation relying solely on monocular RGB information. Consequently, it serves as a lightweight yet viable alternative, eliminating the need for expensive sensors (e.g., LiDAR or depth cameras) in defined environments.

Next, we validate the effectiveness of privileged information by comparing our proposed method against the No-Privilege setup. As shown in Figure \ref{fig:Ablation}, without privileged depth, the policy fails to converge effectively, achieving a reward 20\% less than Proposed. This strongly indicates that privileged information provides the Critic with an accurate spatial understanding. It allows the Critic to form a more stable and accurate value function, which in turn provides a more reliable learning signal for the Actor. 

% Finally, we analyze the Optical Flow Actor. Although NeuFlow \cite{neuralflow} achieves efficient optical flow estimation at a high frame rate, the rewards drops 45\%. This performance degradation is primarily because optical flow inherently discards the rich semantic information present in RGB images. Furthermore, its pairwise estimation approach yields a structural representation of the environment that is significantly weaker than the dense, explicit geometry provided by depth.

 \begin{figure}[h]
	\centering
	\includegraphics[width=1.0\linewidth]{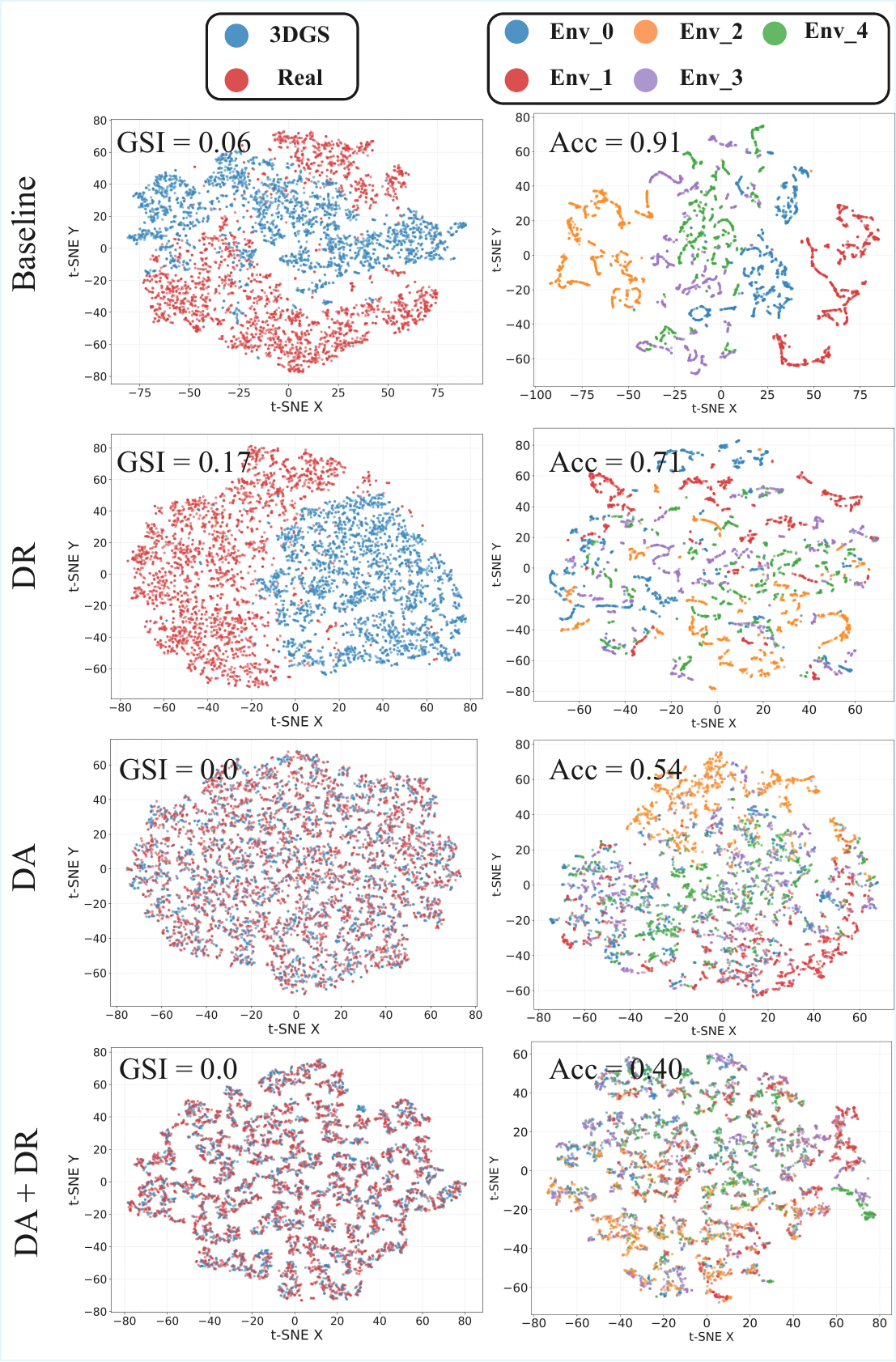}
	\captionsetup{font={small}}
	\caption{
\textbf{t-SNE visualization of the latent feature space.} The left column evaluates domain alignment between 3DGS and Real-world inputs (measured by GSI), while the right column illustrates feature discriminability across different data-augmented environments (measured by Classification Accuracy).
	}
	\vspace{-0.4cm}
	\label{fig:sne}

\end{figure}

\subsection{Visualizing the Sim-to-Real Gap}
To intuitively evaluate the efficacy of our method in bridging the sim-to-real gap and understanding the internal representations, we employ t-SNE to project the high-dimensional latent features extracted by the encoder of the Actor network into a two-dimensional manifold for visualization.
\subsubsection{Alignment between Simulation and Reality}

We first evaluate the sim-to-real alignment by visualizing the feature embeddings of the original 3DGS rendered images against real-world captured images. We employ the Geometric Separability Index (GSI) \cite{alignment} to quantify the degree of region overlap, where a lower GSI indicates better mixing. 

As illustrated in the left column of Fig. \ref{fig:sne}, the Baseline model exhibits a distribution with a GSI of 0.06, indicating a lack of shared feature representation. Interestingly, applying DR leads to a clearer separation between the simulation and real-world clusters, which increases the GSI to 0.17. This phenomenon suggests that the real-world observations remain out-of-distribution to the encoder. In contrast, our DA method successfully aligns the two distributions, making them virtually indistinguishable. This confirms that the encoder has learned to extract domain-invariant features from diverse visual representations. 

As demonstrated in the bottom-left panel of Fig. \ref{fig:sne}, the latent features, trained through a combination of DA and DR, exhibit a highly categorized distribution. Clusters representing similar underlying visual properties become tightly concentrated, while those corresponding to distinct feature types are more spatially segregated. This dual approach not only achieves effective domain alignment between simulation and reality but also enhances the discriminability of different features within the same domain.

\subsubsection{Difference between Generalization and Alignment}

% To further investigate the underlying mechanisms of different training strategies, we visualize the latent vectors of images from five distinct data-augmented environments (Env 0-4). Additionally, we use a simple classifier (Logistic Regression) on these features to quantify their distinguishability via classification accuracy (Acc), where a lower Acc means better alignment. 

% As shown in the right column of Fig. \ref{fig:sne}, the Baseline model produces clearly isolated clusters for each environment with a high classification accuracy of 91\%. This indicates a heavy reliance on environment-specific style features (e.g., texture, lighting) rather than task-relevant semantics. The DR model reduces inter-cluster distances and lowers the accuracy to 71\%. This suggests it achieves robustness by forcing the model to cover a broader range of visual variations, yet still retaining significant domain-specific information. Most remarkably, although our DA method was explicitly trained to align only raw simulation data with real-world data, it achieves the tightest clustering across all five unseen augmented environments, with the classification accuracy dropping sharply to 54\%. Combining DA with DR , the Acc can futher reduce to 40\%. This implies a fundamental difference in mechanism: unlike DR, which relies on expanding the coverage of the feature space, DA forces the model to extract universal geometric and semantic features.

To further investigate the underlying mechanisms of different training strategies, we visualize the latent vectors of images from five distinct data-augmented environments (Env 0-4). Additionally, we use a simple classifier (Logistic Regression) on these features to quantify their distinguishability via classification accuracy (Acc), where a lower Acc means better alignment. 

As shown in the right column of Fig. \ref{fig:sne}, the Baseline model produces clearly isolated clusters for each environment, resulting in a high classification accuracy of 91\%. This indicates that the model heavily relies on environment-specific styles (e.g., texture and lighting) rather than task-relevant semantic information. Introducing DR reduces the distances between clusters and lowers the accuracy to 71\%. This suggests that DR achieves robustness primarily by forcing the model to adapt to a wider range of visual variations. 

Most notably, although our DA method was explicitly trained to align only raw simulation data with real-world data, it achieves tight clustering across all five unseen augmented environments. The classification accuracy drops sharply to 54\%, confirming that the adversarial training effectively removes style-related information. When combining DA with DR, the accuracy is further reduced to 40\%. This comparison highlights a key difference in their mechanisms: unlike DR, which relies on expanding the data coverage to include more styles, DA forces the network to learn universal geometric and semantic features that are invariant to domain shifts.

\subsection{Real-world Flight}

\subsubsection{Ablation Study of Sim-to-Real Methods in Real-World Scenarios}
To further evaluate the transfer performance of various sim-to-real approaches, we conducted validation experiments in real-world scenarios. We established five comparative settings: Baseline, DY, DY+DR, DY+DA, and our proposed Method (combining DR+DA+DY).

As illustrated in Table~\ref{tab:ablation_qualitative}, the Proposed Method demonstrates superior robustness against environmental noise and achieves smooth navigation during sim-to-real transfer. In comparison, the DY+DR method fails to generalize effectively, suggesting that standard visual DR alone is insufficient to cover the complex visual distribution of real-world scenes. 

Furthermore, although the DY+DA method demonstrates the feasibility of navigation, it exhibits a suboptimal success rate. The primary reason is its lack of robustness to environmental interference, such as sensor noise or light condition differences, resulting in frequent collisions with complex obstacles such as foliage. Finally, real-world state estimation, computed via EKF-based fusion of downward optical flow and IMU data, is inherently noisy and uncertain. Thus, without DY, the Baseline fails to bridge the dynamics gap, exhibiting significant instability in the physical environment.

\begin{table}[h]
    \centering
    \caption{ Ablation study on real-world capabilities. We evaluate each variant based on its ability to mitigate the dynamic gap, resist visual interference, and perform visual sim-to-real transfer. }
    \label{tab:ablation_qualitative}

    \resizebox{\columnwidth}{!}{% 强制缩放到栏宽
        \begin{tabular}{l c c c}
            \toprule
            \textbf{Method} & \textbf{Mitigate Dynamic Gap} & \textbf{Visual Robustness} & \textbf{Visual Sim-to-Real} \\
            \midrule
            Baseline & $\times$ & $\times$ & $\times$ \\
            DY & \checkmark & $\times$ & $\times$ \\
            DY + DR & \checkmark & \checkmark & $\times$ \\
            DY + DA & \checkmark & $\times$ & \checkmark \\
            \midrule
            \textbf{DR + DA + DY(Proposed)} & \textbf{\checkmark} & \textbf{\checkmark} & \textbf{\checkmark} \\
            \bottomrule
        \end{tabular}%
        
    } % resizebox 结束
% \vspace{-0.25cm}

\end{table}

\subsubsection{Efficient sim-to-real transfer}

To validate the effectiveness of our sim-to-real approach, as illustrated in Fig. \ref{fig:top}, we conduct comparative experiments in both real-world and simulated environments. Each experimental set consisted of 10 test trajectories. We achieve a 90\% success rate in simulation and an 80\% success rate in the real-world environment. The high success rate demonstrates the successful sim-to-real transfer with our pipeline. The primary cause of failure in the real world was collisions resulting from foliage swaying induced by propeller downwash—a dynamic factor currently unmodeled in the simulation.

\begin{figure}[h]
	\centering
	\includegraphics[width=1.0\linewidth]{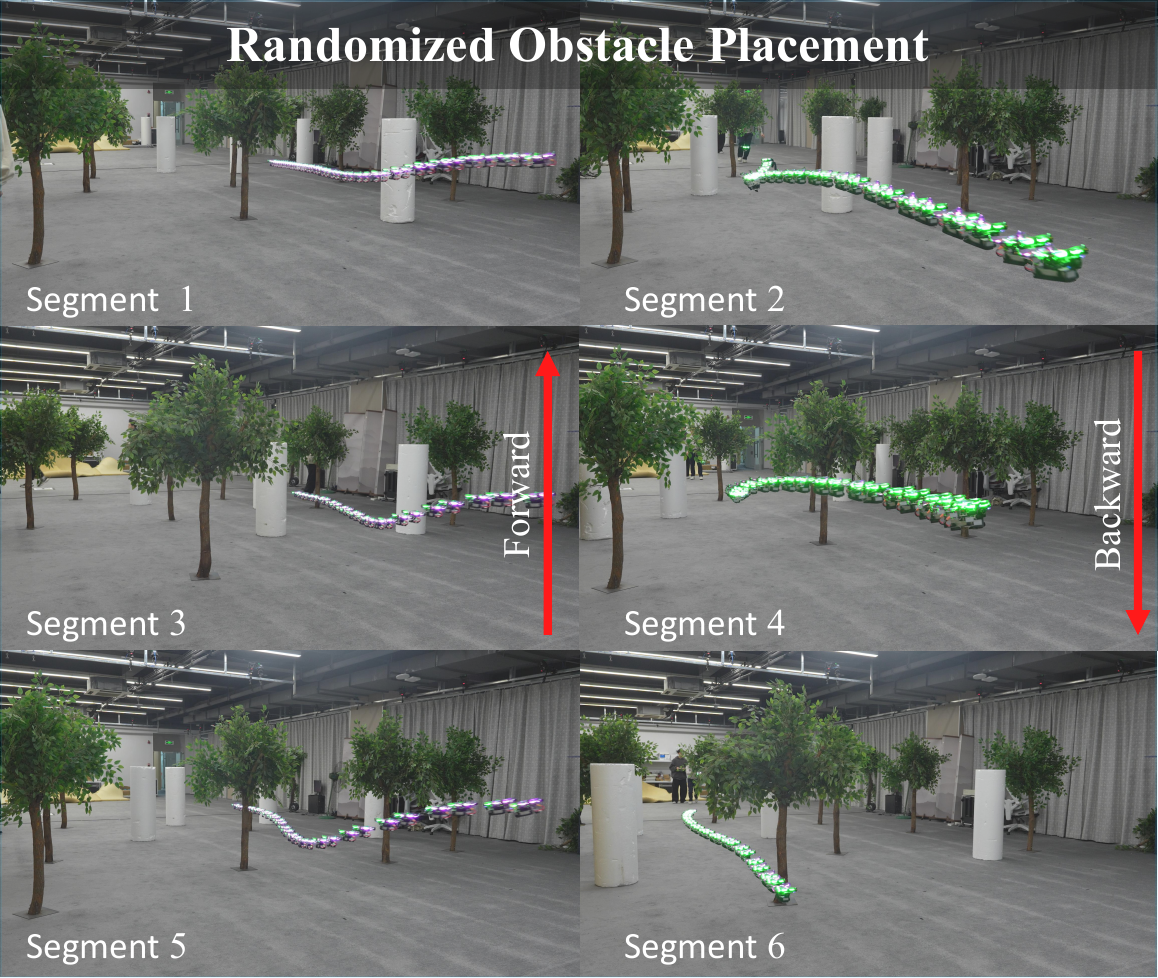}
	\captionsetup{font={small}}
	\caption{
    \textbf{Demonstration of continuous navigation across varying environments.} The mission is divided into six sequential flight segments. Upon reaching the goal, the UAV initiates a return flight and obstacle positions are randomly redistributed. Left: Forward flight trajectory. Right: Return flight trajectory.
	}
	\label{fig:move}
\end{figure}

Furthermore, we evaluate the robustness and zero-shot generalization of our policy through a continuous round-trip navigation task. Crucially, this experiment is conducted as a single, uninterrupted flight session without landing the UAV or resetting the control algorithm. To prevent map memorization, we physically reshuffled the positions of obstacles (including trees and pillars) during the transition between flight legs. As illustrated in Fig. \ref{fig:move}, the left panels depict the forward trajectories, while the right panels show the subsequent return flights, with each flight conducted in a randomly reconfigured environment. Despite the random changes in obstacle distribution, our policy consistently generates smooth, collision-free trajectories. This further validates that our policy successfully extracts generalized visual features, ensuring robust performance across diverse and unseen obstacle configurations.

\subsubsection{Robustness under visual change}

\begin{figure*}[t]
	\centering
	\includegraphics[width=1.0\linewidth]{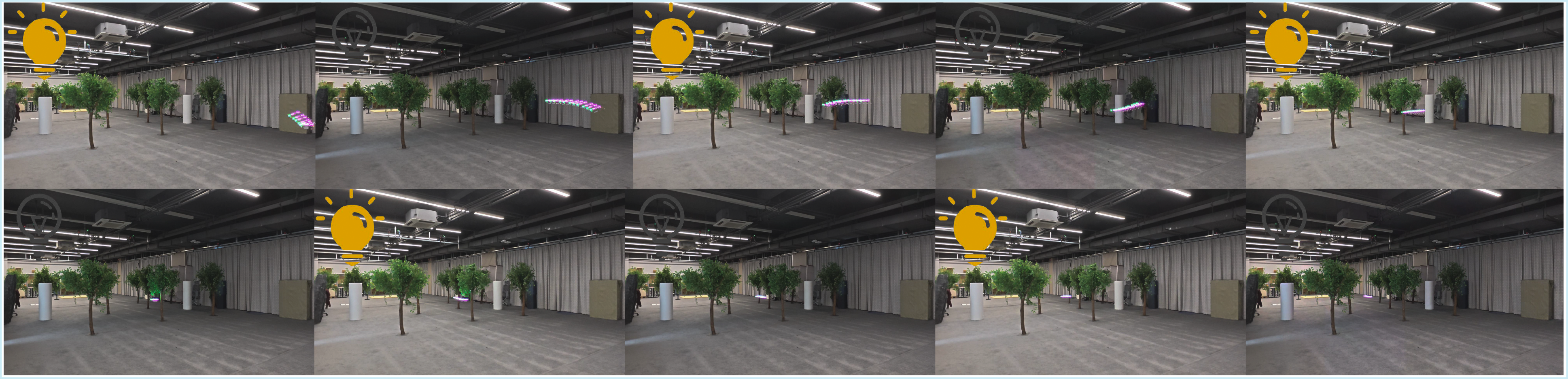}
	\captionsetup{font={small}}
	\caption{
    \textbf{Flight performance under rapidly changing illumination.} The lights are turned on and off rapidly, creating extreme distractions. The resulting smooth trajectory demonstrates the robustness of our policy in such challenging conditions.
	}
	\vspace{-0.5cm}
	\label{fig:change_light}
\end{figure*}

To further evaluate the visual robustness of our policy, we design a flight task characterized by rapidly changing illumination, as shown in Fig. \ref{fig:change_light}. Each frame corresponds to an abrupt shift in lighting conditions. This experiment serves as a stress test for the feature extractor. The results suggest that our framework, driven by DA and DR, effectively extracts semantic geometry despite varying illumination. Instead of relying on low-level pixel intensities that are susceptible to illumination changes, the encoder extracts consistent representations. As a result, the policy exhibits sustained robustness, generating stable control commands despite photometric variations. A more expressive and intuitive experimental demonstration is provided in the supplementary video.

\section{Conclusion and Future Work}

In this paper, we presented a framework combining 3DGS and domain adaptation for monocular RGB-based navigation. We leveraged end-to-end reinforcement learning to extract implicit cues from monocular data. By employing a pruning strategy and parallelized backends, we accelerated the 3DGS rendering speed to 30,000 frames per second. Through experiments in both simulated and real-world environments, we validated that our method achieves effective sim-to-real transfer by incorporating domain adaptation. Notably, the learned policy demonstrated robust navigation capabilities in environments with randomly distributed obstacles and maintained stability under varying illumination conditions.

Despite these results, we acknowledge that training within a limited set of scenes currently limits the policy's adaptability to arbitrary environments. However, our method provides a cost-effective foundation for scaling up simulation data. Future work will focus on utilizing this framework to scale up training across large-scale, diverse 3DGS datasets. By integrating this with VLA \cite{openvla} paradigms, we aim to bridge the gap towards universal navigation policies capable of handling complex, unknown scenarios.

\bibliography{ref}

\end{document}